\documentclass[twoside,leqno,twocolumn]{article}

\usepackage[letterpaper]{geometry}

\usepackage{siamproceedings}

\usepackage[T1]{fontenc}
\usepackage{amsfonts}
\usepackage{booktabs}
\usepackage[skip=2pt]{caption}
\usepackage{graphicx}
\usepackage{epstopdf}
\usepackage{enumitem}
\usepackage{multirow}
\usepackage{algorithmic}
\usepackage{enumitem}
\ifpdf
  \DeclareGraphicsExtensions{.eps,.pdf,.png,.jpg}
\else
  \DeclareGraphicsExtensions{.eps}
\fi

\newsiamremark{remark}{Remark}
\newsiamremark{hypothesis}{Hypothesis}
\crefname{hypothesis}{Hypothesis}{Hypotheses}
\newsiamthm{claim}{Claim}

\usepackage{amsopn}

\begin{document}

\newcommand\relatedversion{}
\renewcommand\relatedversion{\thanks{The full version of the paper can be accessed at \protect\url{https://arxiv.org/abs/0000.00000}}}

\title{\Large When Earth Foundation Models Meet Diffusion: An Application to \\Land Surface Temperature Super-Resolution}
\author{
Yiheng Chen\thanks{University of Alabama, \texttt{ychen226@ua.edu}}
\and
Zihui Ma\thanks{Emory University, \texttt{zihui.ma@emory.edu}}
\and
Peishi Jiang\thanks{University of Alabama, \texttt{peishi.jiang@ua.edu}}
\and
Yilong Dai\thanks{University of Alabama, \texttt{ydai17@ua.edu}}
\and
Qikai Hu\thanks{University of Michigan, \texttt{wsxgshqk@umich.edu}}
\and
Xinyue Ye\thanks{University of Alabama, \texttt{xye10@ua.edu}}
\and
Lingyao Li\thanks{University of South Florida, \texttt{lingyaol@usf.edu}}
\and
Rita Sousa\thanks{New York University, \texttt{rita.sousa@nyu.edu}}
\and
Runlong Yu\thanks{University of Alabama, \texttt{ryu5@ua.edu}.  Runlong Yu is the corresponding author.}
}
\date{}

\maketitle

\fancyfoot[R]{\scriptsize{Copyright \textcopyright\ 2026 by SIAM\\
Unauthorized reproduction of this article is prohibited}}

\begin{abstract}
Land surface temperature (LST) super-resolution is important for environmental monitoring. However, it remains challenging as coarse thermal observations severely underdetermine fine-scale structure. In this paper, we propose Earth Foundation Model-guided Diffusion (EFDiff), a novel framework for super-resolution under extreme spatial degradation. EFDiff uses the Prithvi-EO-2.0 Earth foundation model to encode high-resolution multispectral reflectance into geospatial embeddings, which are injected into the denoising network via cross-attention to guide fine-scale reconstruction from highly degraded observations. We study two variants, EFDiff-$\epsilon$ and EFDiff-$x_0$, which offer complementary trade-offs between perceptual realism and pixel-level fidelity.
We evaluate EFDiff under an extreme $32\times$ scale gap using a globally diverse benchmark comprising 242,416 co-registered Landsat thermal-reflectance patches. Results show that EFDiff consistently outperforms baseline methods and that cross-attention conditioning by EFM is more effective than HLS channel concatenation. Although we present EFDiff in the context of LST super-resolution, the framework is broadly applicable to remote sensing problems in which pretrained geospatial representations can guide generative reconstruction.
\end{abstract}

\section{Introduction.}
\label{sec:intro}

Land surface temperature (LST) regulates the radiative and turbulent exchange of energy between the Earth's surface and the atmosphere and is central to applications such as urban heat island monitoring, evapotranspiration estimation, drought assessment, and ecosystem health analysis \cite{li2013satellite,srivanit2012thermal}. Satellite thermal infrared observations provide spatially continuous LST retrievals, but operational products are constrained by a long-standing trade-off between spatial and temporal resolution \cite{zhan2013disaggregation}: MODIS offers near-daily global coverage at approximately 1\,km resolution, whereas Landsat provides much finer thermal observations at approximately 100\,m resolution but with a 16-day revisit cycle \cite{hulley2021validation,wan1999modis}. Many downstream applications, however, require both timely coverage and fine spatial detail, making the reconstruction of high-resolution thermal fields from coarse observations an important problem for environmental monitoring.

Reconstructing fine-scale LST from coarse observations remains fundamentally challenging. A coarse thermal pixel may aggregate multiple land-cover types, surface materials, moisture conditions, and microclimatic regimes while discarding the fine-scale boundaries that govern local thermal contrast, making extreme LST super-resolution a severely ill-posed geospatial inverse problem. A large body of work has studied LST sharpening and super-resolution, ranging from regression-based approaches to modern deep neural networks \cite{Pu2023review,kustas2003estimating,agam2007vegetation}. However, three structural challenges remain insufficiently addressed. First, most existing methods are fundamentally deterministic, so under large-scale gaps they tend to minimize pixel-wise error by predicting conditional means, yielding blurred spatial transitions and attenuated local extremes. Second, most methods are developed and evaluated within a single study region or limited geographic domain, which limits the transferability of learned spectral--thermal relationships across climates, biomes, and land-cover regimes. This limitation is especially problematic for LST, whose fine-scale patterns are shaped by geographically variable interactions among vegetation, built surfaces, water bodies, terrain, and moisture. Third, the conditioning signals used by current methods are typically hand-crafted variables, physical priors, or raw multispectral bands. Although these inputs provide useful low-level cues, they do not explicitly encode the higher-level geospatial semantics needed to disambiguate subpixel mixtures and reconstruct sharp thermal contrasts, particularly in heterogeneous scenes. These limitations become especially restrictive in extreme settings such as $32\times$ super-resolution, where a single coarse pixel may summarize more than $10^3$ target pixels.

Recent progress in generative modeling and foundation models suggests a promising direction. Diffusion models are well-suited to extreme super-resolution because they can represent multiple plausible high-resolution solutions \cite{ho2020denoising,saharia2022image,yue2023resshift}. Earth foundation models (EFMs), meanwhile, provide transferable spectral--spatial representations of land-surface structure from large-scale satellite pretraining \cite{cong2022satmae,szwarcman2025prithvi}. The two are naturally complementary: diffusion models uncertainty in reconstruction, while EFMs provide semantically rich geospatial context. However, these directions have largely evolved in isolation. Existing diffusion-based LST methods still condition the denoising process on hand-crafted geophysical variables or raw spectral inputs \cite{zhang2025pgdm,chen2024super,wang2025information}, leaving the model to infer high-level land-surface semantics from relatively limited task-specific supervision. Conversely, EFMs have been used primarily for discriminative remote-sensing tasks such as classification and segmentation \cite{cong2022satmae,szwarcman2025prithvi}. What remains missing is a unified framework that couples generative reconstruction with semantically rich, globally pretrained geospatial representations.

To address this gap, we propose Earth Foundation Model-guided Diffusion (EFDiff), a framework for super-resolution under extreme spatial degradation. Our motivation is that, in extreme LST super-resolution, the key challenge is not only to generate plausible fine-scale detail, but to resolve ambiguity using semantically meaningful land-surface context. Raw multispectral inputs provide low-level cues, but they do not explicitly encode the higher-level geospatial semantics needed for this purpose. EFDiff therefore uses a pretrained Earth foundation model as a conditioning encoder for diffusion. Specifically, Prithvi-EO-2.0 \cite{szwarcman2025prithvi} encodes high-resolution multispectral reflectance into geospatial embeddings, which are injected into the denoising network through cross-attention. We further study two variants, EFDiff-$\epsilon$ and EFDiff-$x_0$, to examine the trade-off between perceptual realism and pixel-level fidelity under the same conditioning design. We evaluate EFDiff on extreme LST super-resolution using a globally diverse benchmark of 242{,}416 co-registered Landsat thermal--reflectance patches under a unified $32\times$ protocol. Results show that EFM cross-attention conditioning is more effective than HLS channel concatenation, especially in heterogeneous scenes, and that the two variants offer complementary strengths.
Our contributions are summarized as follows:
\begin{itemize}[nosep,leftmargin=*]
  \item We propose EFDiff, a novel framework that integrates Earth foundation models and diffusion-based generative modeling for LST super-resolution, demonstrating how pretrained geospatial representations can serve as guidance for generative reconstruction.
  \item We study two complementary formulations, EFDiff-$\epsilon$ and EFDiff-$x_0$, showing how pretrained geospatial conditioning interacts with diffusion objectives to offer different trade-offs between perceptual realism and pixel-level fidelity.
  \item We introduce a globally diverse benchmark for LST super-resolution and show that EFM cross-attention conditioning outperforms HLS channel concatenation, especially in spatially heterogeneous scenes.
\end{itemize}

 \section{Related Work.}
\label{sec:related}

LST super-resolution has long been approached by exploiting relationships between temperature and surface properties observed at finer resolution. Early methods such as DisTrad \cite{kustas2003estimating} and TsHARP \cite{agam2007vegetation} regress coarse LST against vegetation-related predictors and refine the result with residual correction, while later extensions incorporate albedo, emissivity, elevation, and nonlinear regressors such as random forests \cite{hutengs2016downscaling}. More recently, physically motivated formulations based on surface energy balance have been introduced to move beyond purely statistical predictors \cite{firozjaei2024novel}. However, these methods remain most effective at moderate scale factors ($2$--$4\times$) and rely on relationships that are difficult to preserve under large spatial gaps and heterogeneous land cover \cite{Pu2023review}. Deep models improve flexibility by learning richer mappings from multisource inputs \cite{xu2025downscaling,dai2025mocolsk}, but existing studies are still largely region-specific. Diffusion-based LST methods further improve realism by modeling uncertainty in fine-scale reconstruction, yet they are typically conditioned on hand-crafted geophysical variables, auxiliary maps, or raw spectral bands \cite{zhang2025pgdm,chen2024super,wang2025information}, and are mostly evaluated on regional datasets at scale factors no larger than $20\times$. This leaves a clear task-level gap: extreme, globally diverse LST super-resolution still lacks a generative framework that can leverage stronger pretrained geospatial priors.

This gap is closely tied to the broader design space of diffusion-based super-resolution. One key question is how the conditioning signal enters the denoising network. SR3~\cite{saharia2022image} established the standard practice of HLS channel concatenation of the low-resolution input with the noisy sample, which is simple and effective but largely confines conditioning to pixel-aligned, low-level cues. Latent diffusion models (LDMs)~\cite{rombach2022high} introduced a more expressive alternative by injecting external embeddings through cross-attention, allowing generation to be guided by richer pretrained representations. A second question is how the reverse process is parameterized at inference time. Standard DDPMs~\cite{ho2020denoising} require many denoising steps, whereas DDIM~\cite{song2020denoising} enables faster deterministic sampling, and ResShift~\cite{yue2023resshift} further accelerates super-resolution through a residual-shifting formulation. These choices define how diffusion models trade off fidelity, realism, and efficiency.

Earth foundation models provide the missing ingredient. Recent self-supervised models such as SatMAE \cite{cong2022satmae}, DOFA \cite{xiong2024neural}, SkySense \cite{guo2024skysense}, AlphaEarth~\cite{brown2025alphaearth}, Clay \cite{clay_foundation_model_2024}, and Prithvi-EO-2.0 \cite{szwarcman2025prithvi} show that large-scale pretraining on remote sensing imagery yields transferable geospatial representations that outperform conventional visual backbones across benchmark tasks, including GEO-Bench \cite{lacoste2023geo}. However, these models have been used primarily for discriminative tasks such as classification, segmentation, and change detection, not as conditioning encoders for generative reconstruction. That distinction is central to our contribution. The novelty is not simply using diffusion, nor simply using EFMs, but using frozen, spatially structured pretrained EFM embeddings as conditioning signals for generative super-resolution. Concretely, our method injects EFM embeddings via cross-attention into the diffusion denoiser, moving beyond the hand-crafted variables and HLS channel concatenation used in prior LST diffusion work \cite{zhang2025pgdm,chen2024super,wang2025information}. In doing so, we target a substantially harder setting: $32\times$ LST super-resolution on a globally diverse benchmark spanning more than $3,000$ tiles and 242k patches.

\section{Problem Definition.}
\label{sec:problem}

We study conditional super-resolution for LST under an extreme spatial scale gap. Let $Y \in \mathbb{R}^{H \times W}$ denote the target high-resolution (HR) LST field at pixel spacing $\delta_{\mathrm{HR}}$, and let $X \in \mathbb{R}^{h \times w}$ denote the corresponding low-resolution (LR) thermal observation at spacing $\delta_{\mathrm{LR}}$, with scale factor $s = \delta_{\mathrm{LR}} / \delta_{\mathrm{HR}}$. We also assume access to co-registered multispectral reflectance $S \in \mathbb{R}^{C \times H \times W}$ on the HR grid, which provides auxiliary land-surface information aligned with $Y$. In our setting, $s = 32$: each coarse thermal pixel at 960\,m corresponds to $32 \times 32 = 1{,}024$ target pixels at 30\,m.

Recovering $Y$ from $X$ is severely ill-posed. A single coarse thermal pixel may aggregate multiple land-cover types, surface materials, moisture conditions, and microclimatic regimes while discarding the fine-scale boundaries that govern local thermal contrast. As a result, many distinct HR fields can be consistent with the same coarse observation, making the conditional distribution $p(Y \mid X)$ highly multimodal. This also explains why deterministic predictors tend to produce blurred outputs: under large spatial gaps, minimizing pixel-wise error encourages regression toward a conditional mean rather than recovery of sharp local structure.

To make the problem more tractable, we first map the coarse observation to the HR grid using an upsampling operator,
$
\tilde{X} = \mathcal{U}(X) \in \mathbb{R}^{H \times W},
$
and reformulate the task in residual space:
$
R = Y - \tilde{X}.
$
The goal is then to recover a fine-scale residual field that refines the coarse thermal structure already present in $\tilde{X}$. This yields the reconstruction
$
\hat{Y} = \tilde{X} + \hat{R}.
$
In our instantiated model, the auxiliary reflectance $S$ is further encoded into geospatial embeddings $Z = E_{\phi}(S)$, which provide semantic spatial context for disambiguating subpixel mixtures and thermal boundaries. The learning problem is thus to model the conditional distribution
$
p(R \mid \tilde{X}, Z).
$ This residual-space conditional formulation motivates the diffusion-based generative framework.

\section{Method.}
\label{sec:method}

\begin{figure*}[t]
  \centering
  \includegraphics[width=\textwidth]{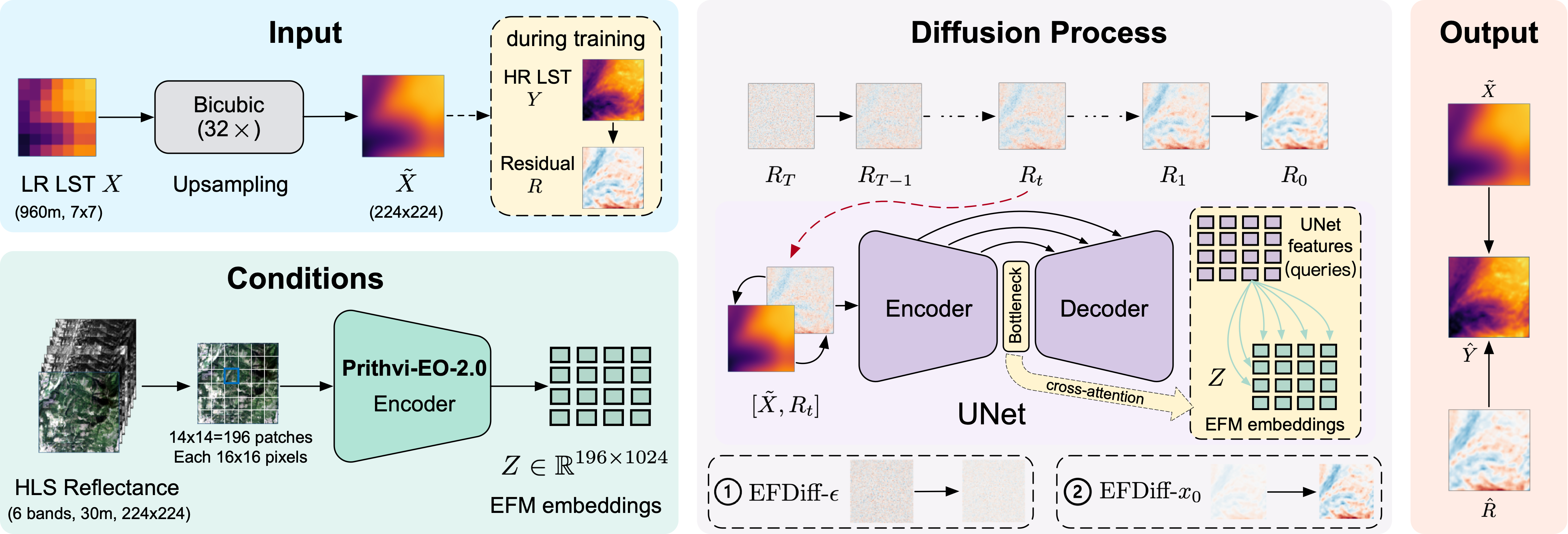}
  \vspace{0.01cm}
  \caption{Overall framework of EFDiff.}
  \vspace{-0.3cm}
  \label{fig:framework}
\end{figure*}

As illustrated in Fig.~\ref{fig:framework}, the framework takes the upsampled coarse thermal field $\tilde{X}$ and the co-registered HLS reflectance as inputs and generates the predicted residual $\hat{R}$ through an iterative denoising process in residual space, guided by high-level land-surface features. We first introduce the foundation model that encodes HLS reflectance into a set of geospatial embeddings, then describe the denoising network that fuses these embeddings with the thermal signal via cross-attention, and finally present the two diffusion formulations used for training and inference.

\subsection{EFM-Guided Conditioning.}
\label{sec:foundation}

The auxiliary reflectance $S$ provides rich spatial information at the target resolution, but how this information is represented matters. A straightforward approach would be to feed the raw HLS bands directly to the denoising network as additional input channels. However, raw pixel values carry primarily low-level spectral information and lack the semantic understanding needed to distinguish, for example, a vegetation--urban boundary from a water--soil transition---distinctions that strongly shape the local thermal landscape. Learning such high-level spatial reasoning from the relatively small LST training set alone would be both data-inefficient and prone to overfitting.

We leverage a pretrained EFM to obtain a semantically rich encoding of $S$. Specifically, we adopt the Prithvi-EO-2.0 encoder~\cite{szwarcman2025prithvi}, a ViT-Large model~\cite{dosovitskiy2020image} pretrained with the masked autoencoder (MAE) objective~\cite{he2022masked} on a global corpus of HLS imagery at 30\,m resolution. Given a six-band HLS input, the encoder tokenizes it into non-overlapping $16 \times 16$ patches, processes the resulting sequence through transformer blocks, and outputs a set of geospatial embeddings $Z \in \mathbb{R}^{N \times d}$, where $N = (224/16)^2 = 196$ and $d = 1024$.

We use the encoder as a frozen feature extractor throughout training: its weights are fixed, and gradients do not propagate into it. This design choice has three advantages. First, freezing avoids catastrophic forgetting of the pretrained representations. Second, it decouples the EFM's parameter count from the training cost, since no optimizer state is maintained for the encoder. Third, the resulting embeddings provide a stable, globally informed spatial context: because the encoder has seen millions of diverse land surfaces during pretraining, it captures vegetation boundaries, urban fabric, water bodies, and terrain texture that are strongly correlated with fine-scale thermal structure. These embeddings serve as the conditioning tokens and are injected into the denoising network.

\subsection{Denoising Network.}
\label{sec:network}

The denoising network $D_\theta$ takes as input a noisy residual state $R_t$ concatenated with the upsampled coarse field $\tilde{X}$, together with diffusion timestep $t$, and predicts a quantity that is used to recover the clean residual during reverse diffusion. We adopt a UNet encoder--decoder architecture and condition it on the geospatial embeddings $Z$ via cross-attention layers.
The encoder consists of a sequence of residual blocks with optional self-attention, interleaved with spatial downsampling. The decoder mirrors this structure with upsampling layers and skip connections from the encoder. At the network input, $\tilde{X}$ is channel-concatenated with $R_t$ so that the coarse thermal structure is available to guide denoising from the earliest layers. The diffusion timestep $t$ is encoded via a sinusoidal positional embedding, projected through a two-layer MLP, and injected into each residual block as an additive bias on the feature maps.

Following the cross-attention mechanism introduced for text-conditioned image generation~\cite{rombach2022high}, we augment each self-attention site in the decoder and bottleneck with a cross-attention layer conditioned on the geospatial embeddings $Z$. In this design, the upsampled coarse field $\tilde{X}$ is used only through channel concatenation at the network input, while the geospatial embeddings $Z$ provide the external context for attention. Concretely, the queries are derived from the current UNet feature map $F \in \mathbb{R}^{C \times H' \times W'}$, whereas the keys and values come from the embeddings $Z$:
\begin{equation}
  \label{eq:cross_attn}
  \mathrm{CrossAttn}(F, Z) = \mathrm{softmax}\!\Bigl(\frac{Q^{(F)}\, {K^{(Z)}}^\top}{\sqrt{d_h}}\Bigr)\, V^{(Z)},
\end{equation}
where $Q^{(F)} = W_Q\, F$, $K^{(Z)} = W_K\, Z$, $V^{(Z)} = W_V\, Z$, and $d_h = C / n_{\mathrm{heads}}$. The resulting attention output is added back to $F$ through a residual connection after layer normalization and projection. Each decoder block follows the sequence: residual block $\to$ self-attention $\to$ cross-attention.

The encoder processes only the thermal input ($R_t$ and $\tilde{X}$), while the HLS context is fused only in the decoder and bottleneck. The intuition is that the encoder should first extract resolution-invariant thermal features without being biased by the HR spatial structure in $Z$, which could otherwise allow the network to bypass learning from the thermal signal. The spectral--spatial context is most valuable during upsampling, where the network reconstructs fine-grained detail that the coarse thermal input cannot provide.

\subsection{Diffusion Training and Inference.}
\label{sec:diffusion}

We train the denoising network $D_\theta$ under the diffusion framework~\cite{ho2020denoising,song2020score}, in which a forward process gradually corrupts the clean residual $R$ with noise and the network learns to reverse this corruption. At every timestep, the network receives $\tilde{X}$ as a concatenated input and is conditioned on $Z$ via cross-attention. We study two formulations under this shared conditioning design, denoted as EFDiff-$\epsilon$ and EFDiff-$x_0$. They share the same network architecture but differ in their forward process and prediction target.

\noindent \textbf{EFDiff-$\epsilon$.}
The first variant follows the DDPM framework~\cite{ho2020denoising} adapted for conditional super-resolution~\cite{saharia2022image}. A variance-preserving forward process corrupts $R$ over $T$ timesteps by progressively adding Gaussian noise until the signal is fully destroyed. The network is trained to predict the noise $\epsilon$ added at each step:
\begin{equation}
  \label{eq:eps_loss}
  \mathcal{L}_{\epsilon} = \mathbb{E}_{R,\, \epsilon,\, t}\, \bigl\lVert \epsilon - D_\theta(R_t,\, \tilde{X},\, Z,\, t) \bigr\rVert_1,
\end{equation}
where $\epsilon \sim \mathcal{N}(0, \mathbf{I})$ and $t$ is sampled uniformly. At inference, sampling starts from pure Gaussian noise and iteratively removes the predicted noise to recover a residual estimate $\hat{R}$. The deterministic DDIM sampler~\cite{song2020denoising} can also be applied to reduce the number of function evaluations without retraining.

\noindent \textbf{EFDiff-$x_0$.}
The second variant adopts the transition-based forward process of ResShift~\cite{yue2023resshift}, in which the noisy state interpolates between the clean residual and scaled Gaussian noise rather than progressively destroying the signal. The network directly predicts the clean residual target $R$ at each step:
\begin{equation}
  \label{eq:x0_loss}
  \mathcal{L}_{x_0} = \mathbb{E}_{R,\, \epsilon,\, t}\, \bigl\lVert R - D_\theta(R_t,\, \tilde{X},\, Z,\, t) \bigr\rVert_2^2.
\end{equation}
Because this forward process transitions more rapidly toward noise than the variance-preserving schedule, the formulation naturally operates with far fewer timesteps while maintaining sample quality.

These two variants offer complementary strengths within the same conditioning framework. EFDiff-$\epsilon$ excels at generating perceptually realistic spatial textures through its longer stochastic sampling trajectory, while EFDiff-$x_0$ delivers superior pixel-level fidelity at a fraction of the computational cost. This dual-variant design enables practitioners to select the version best suited to their application requirements.

\section{Experiments.}
\label{sec:experiments}

We organize the empirical study around four research questions:
\begin{itemize}[leftmargin=1.0em, itemsep=1pt, topsep=1pt, parsep=0pt, partopsep=0pt]
  \item \textbf{RQ1:} Does EFDiff outperform baselines for $32\times$ LST super-resolution on a globally diverse benchmark?
    
  \item \textbf{RQ2:} What trade-off do EFDiff-$\epsilon$ and EFDiff-$x_0$ induce between distortion, perceptual quality, and spatial error structure?
  \item \textbf{RQ3:} Do the proposed models better recover fine-scale thermal structure in qualitative case studies?
  \item \textbf{RQ4:} Are the gains explained by EFM cross-attention conditioning, and how sensitive are the two formulations to sampling steps?
\end{itemize}



\subsection{Dataset.}
\label{sec:dataset}

\begin{figure}[t]
  \centering
  \includegraphics[width=\columnwidth]{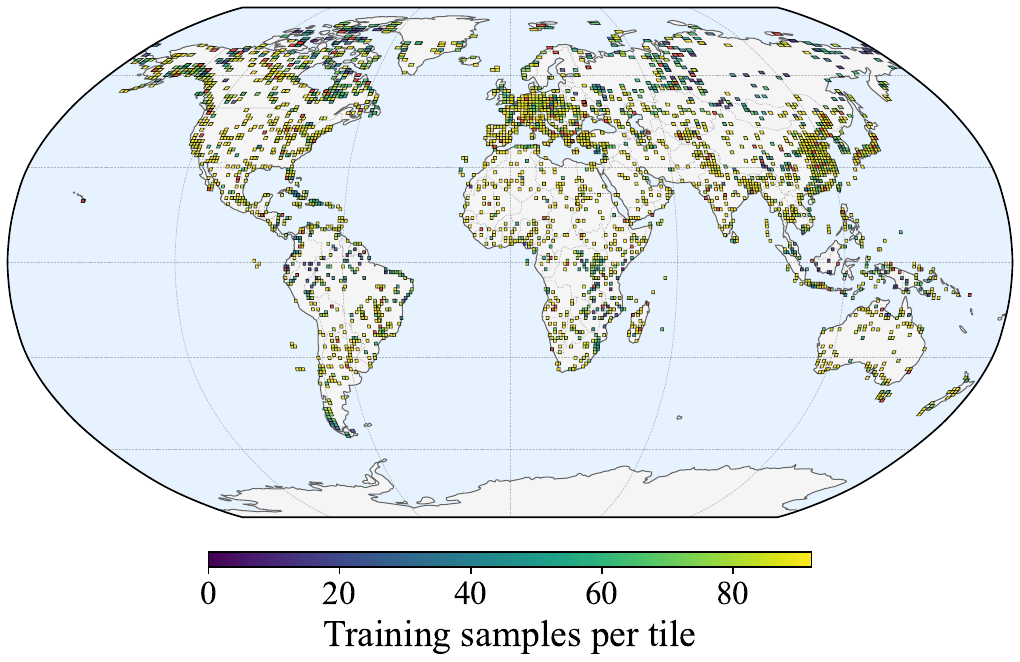}
  \caption{Global distribution of training and test tiles. Color indicates the number of patches per tile (viridis colormap); test tiles are shown in red.}
  \vspace{-0.5cm}
  \label{fig:data_distribution}
\end{figure}

We construct a globally diverse benchmark from NASA's Harmonized Landsat Sentinel-2 (HLS) archive~\cite{claverie2018harmonized}. We use six-band reflectance from the Landsat-only \texttt{HLSL30} product and pair each sample with the corresponding Landsat Collection~2 Level-2 \texttt{ST\_B10} land surface temperature (LST) observation~\cite{cook2014development}, retaining only matches within $\pm$1\,day. Following the stratified tile sampling strategy of Prithvi-EO-2.0~\cite{szwarcman2025prithvi}, the dataset spans 3{,}094 MGRS tiles (2{,}943 training, 151 test) and contains 242{,}416 co-registered $256\!\times\!256$ patches (230{,}659 training, 11{,}757 test) collected from 2014 to 2026. Figure~\ref{fig:data_distribution} shows the global distribution of the sampled tiles.
Each sample consists of an HR LST field $Y$ at 30\,m and a co-registered six-band reflectance image $S$. During training, we randomly crop $224\!\times\!224$ sub-patches; at inference, we use a deterministic center crop. The LR thermal input is synthesized under Wald's reduced-resolution protocol~\cite{wald1997fusion}: the HR target is degraded to a $7\!\times\!7$ coarse field and then bicubically upsampled back to $224\!\times\!224$ to form $\tilde{X}$. We predict the residual $R = Y - \tilde{X}$, with $R$ standardized to $[-1,1]$ and reflectance normalized to $[0,1]$. Full details on dataset construction, filtering, leakage prevention, and normalization are provided in Appendix~\ref{sec:appendix_data}.

\begin{table*}[t]
\centering
\caption{Overall performance comparison. Bold indicates the best result for each metric within each group.}
\small
\setlength{\tabcolsep}{3pt}
\begin{tabular}{@{}c|cccc|cccc|cccc@{}}
\toprule
 & \multicolumn{4}{c|}{\textbf{Forest}} & \multicolumn{4}{c|}{\textbf{Low Vegetation}} & \multicolumn{4}{c}{\textbf{Water}} \\
\textbf{Model} & RMSE$\downarrow$ & SSIM$\uparrow$ & LPIPS$\downarrow$ & FID$\downarrow$
               & RMSE$\downarrow$ & SSIM$\uparrow$ & LPIPS$\downarrow$ & FID$\downarrow$
               & RMSE$\downarrow$ & SSIM$\uparrow$ & LPIPS$\downarrow$ & FID$\downarrow$ \\
\midrule
Bicubic            & 1.1812 & 0.4407 & 0.8209 & 371.70   & 1.4422 & 0.4091 & 0.8321 & 375.32   & 0.6860 & 0.4780 & 0.7823 & 330.92 \\
SRGAN              & 0.6468 & 0.7305 & 0.3942 & 45.93   & 0.7724 & 0.7097 & 0.4024 & 42.63   & 0.4106 & 0.5845 & 0.5903 & 120.45 \\
HAT                & 0.5998 & 0.7393 & 0.3954 & 43.11   & 0.7217 & 0.7210 & 0.3974 & 39.29   & 0.3768 & 0.5910 & 0.5950 & 108.79 \\
SwinIR             & 0.6186 & 0.7322 & 0.3899 & 39.88   & 0.7407 & 0.7128 & 0.3956 & 37.76   & 0.3852 & 0.5864 & 0.5962 & 106.86 \\
EDSR               & 0.6615 & 0.7307 & 0.3680 & 35.69   & 0.7854 & 0.7118 & 0.3858 & 36.02   & 0.4218 & 0.5821 & 0.5717 & 104.64 \\
RCAN               & 0.5336 & 0.7729 & 0.3554 & 39.61   & 0.6480 & 0.7558 & 0.3693 & 37.41   & 0.3381 & 0.6073 & 0.5826 & 112.86 \\
SR3                & 0.6826 & 0.6700 & 0.2668 & 12.13   & 0.8527 & 0.6399 & 0.2675 & 10.43   & 0.4202 & 0.4866 & 0.4605 & 36.73 \\
ResShift           & 0.5619 & 0.7462 & 0.2568 & 18.95   & 0.6882 & 0.7249 & 0.2807 & 19.98   & 0.3421 & 0.5820 & 0.2875 & 41.82 \\
EFM Regression     & 0.5378 & 0.7520 & 0.4441 & 77.88   & 0.6433 & 0.7400 & 0.4375 & 68.83   & 0.3378 & 0.6007 & 0.6083 & 129.49 \\
\midrule
EFDiff-$\epsilon$ & 0.6224 & 0.6939 & \textbf{0.2225} & \textbf{9.80}   & 0.7555 & 0.6761 & \textbf{0.2243} & \textbf{7.10}   & 0.3821 & 0.5182 & \textbf{0.2218} & \textbf{22.49} \\
EFDiff-$x_0$        & \textbf{0.4660} & \textbf{0.7912} & 0.3662 & 51.54   & \textbf{0.5629} & \textbf{0.7794} & 0.3766 & 47.11   & \textbf{0.2892} & \textbf{0.6199} & 0.5644 & 116.86 \\
\bottomrule
\end{tabular}
\label{tab:per_lulc}
\end{table*}

\begin{figure*}[t]
  \centering
  \vspace{-0.2cm}
  \includegraphics[width=\textwidth]{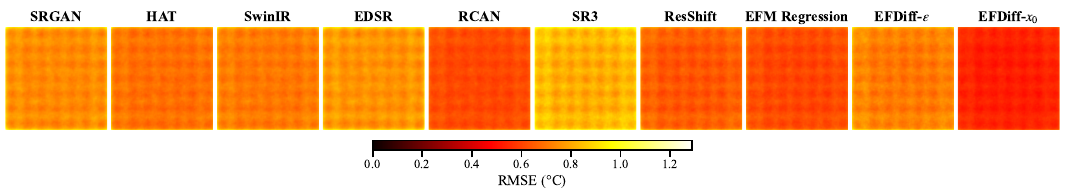}
  \caption{Per-pixel RMSE maps aggregated over the test set.}
  \vspace{-0.3cm}
  \label{fig:spatial_rmse_map}
\end{figure*}

\subsection{Experimental Setup.}
\label{sec:implementation}

All models operate on $224\!\times\!224$ patches under the $32\times$ super-resolution protocol and predict the residual $R = Y - \tilde{X}$. The proposed EFDiff-$\epsilon$ and EFDiff-$x_0$ models share the architecture described in Sec.~\ref{sec:network}: both use the same UNet backbone and a frozen Prithvi-EO-2.0-300M encoder for EFM cross-attention conditioning, but differ in diffusion formulation, training objective, and sampling schedule. 
We compare against nine baselines spanning residual CNNs (EDSR, RCAN), perceptual/adversarial super-resolution (SRGAN), attention-based restoration (SwinIR, HAT), diffusion models (SR3, ResShift), and an EFM-guided regression baseline. All non-EFM baselines receive the same 7-channel input formed by concatenating $\tilde{X}$ with the six HLS bands. Diffusion baselines use the same UNet backbone as the proposed models but replace EFM cross-attention conditioning with HLS channel concatenation.

All models are trained with Adam and evaluated using EMA weights. We report two complementary metric groups: \emph{distortion} metrics (RMSE in ${}^{\circ}$C and SSIM~\cite{wang2004image}) and \emph{perceptual} metrics (LPIPS~\cite{zhang2018unreasonable} and FID~\cite{heusel2017gans}). This distinction is important for extreme $32\times$ super-resolution, where low pixel-wise error does not necessarily imply realistic fine-scale thermal structure~\cite{blau2018perception}. To reveal differences across surface types, all results are reported separately for three land-use/land-cover groups derived from the Copernicus 100\,m land-cover product~\cite{buchhorn2020copernicus}: \emph{Forest}, \emph{Low Vegetation}, and \emph{Water}. Additional implementation and training details are provided in Appendix~\ref{sec:appendix_setup}.

\begin{figure*}[t]
  \centering
  \includegraphics[width=\textwidth]{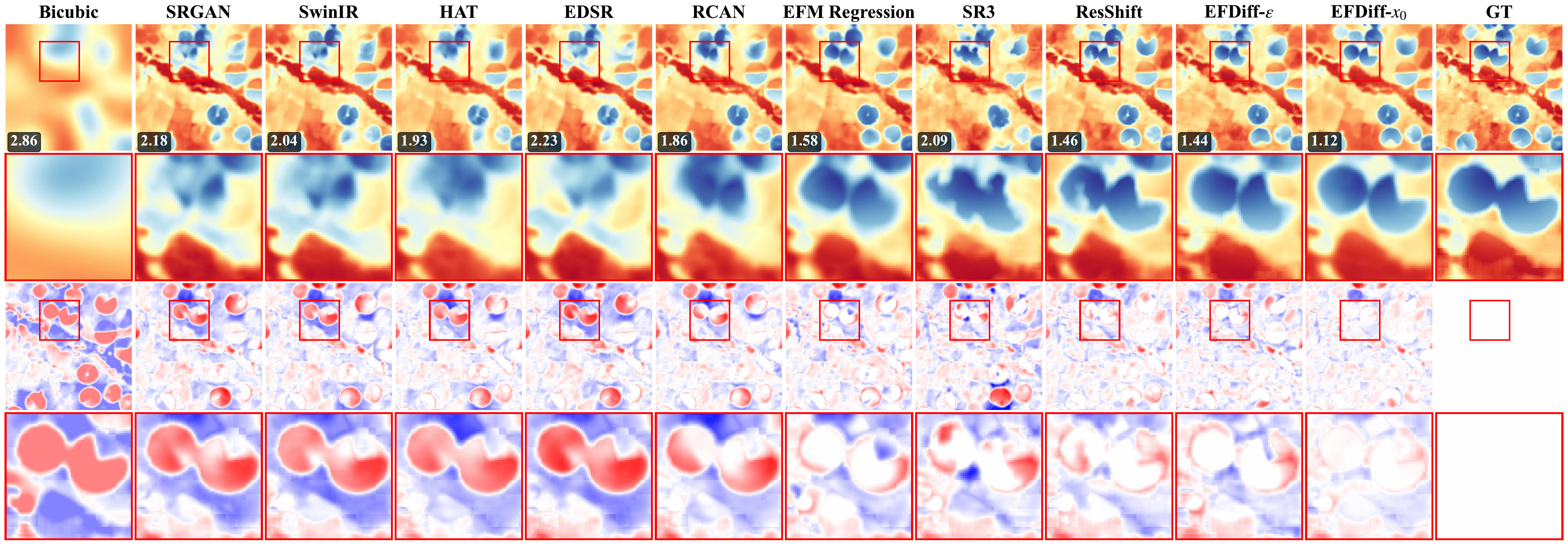}
  \caption{Qualitative comparison on a representative test patch. From top to bottom: reconstructed field $\hat{Y}$, zoomed view of the red-boxed region, error map $\hat{Y}-Y$, and zoomed error map. RMSE (${}^{\circ}$C) is shown in the top-left corner of each prediction panel.}
  \vspace{-0.3cm}
  \label{fig:visual_comparison}
\end{figure*}

\subsection{Model Performance.}
\label{sec:performance}

Table~\ref{tab:per_lulc} reports per-LULC results. Bold indicates the best result for each metric within each group. We discuss the key findings organized by model family.

\paragraph{Deterministic and adversarial baselines.}
Among the deterministic methods, RCAN consistently achieves the lowest RMSE (0.53--0.65$\,{}^{\circ}$C across LULC groups), indicating that channel attention effectively selects informative spectral bands. The other CNN and Transformer baselines (EDSR, SwinIR, and HAT) achieve comparable RMSE but remain less accurate overall. SRGAN does not improve distortion and yields elevated FID, suggesting that adversarial training alone is insufficient to recover realistic thermal textures. The EFM-guided regression baseline matches RCAN in RMSE but produces the worst perceptual scores among all learned models (FID $>$ 68), consistent with single-pass prediction collapsing toward a blurred conditional mean.

\paragraph{Diffusion baselines and proposed models.}
SR3 and ResShift, both conditioned via HLS channel concatenation, already demonstrate the advantage of generative modeling: SR3 achieves the lowest LPIPS among non-EFM baselines ($\sim$0.27) and competitive FID, while ResShift offers a stronger distortion--perception balance. However, both are consistently outperformed by their EFM-conditioned counterparts. EFDiff-$x_0$ achieves the best distortion performance overall (RMSE 0.29--0.56$\,{}^{\circ}$C and the highest SSIM in every category), surpassing all deterministic baselines including RCAN. EFDiff-$\epsilon$ achieves the best perceptual scores by a wide margin (LPIPS $\sim$0.22, FID below 10 on land categories, corresponding to roughly a $4\times$ improvement over the best baseline FID), at the cost of moderately higher pixel-wise error. This behavior is consistent with the formulation: the variance-preserving forward process distributes the signal over $1,000$ timesteps, and stochastic reverse sampling introduces pixel-level variation even when the generated texture is distributionally realistic.

Across all methods, perceptual metrics are notably worse on Water (FID $>$ 100 for most models, versus $<$ 50 on land). This suggests a conditioning gap: water surfaces are spectrally uniform yet thermally heterogeneous, so HLS reflectance offers little discriminative structure, and the Prithvi-EO-2.0 encoder may provide less informative context over open water than over land. Even so, EFDiff-$x_0$ reduces Water RMSE to 0.29$\,{}^{\circ}$C, a 16\% improvement over the best non-EFM baseline.

\paragraph{Spatial error structure.}
We compute a per-pixel RMSE map for each model to reveal position-dependent biases: at spatial location $(i,j)$ in the $224\!\times\!224$ grid, the squared error is averaged over all test patches and the square root is then taken. Figure~\ref{fig:spatial_rmse_map} shows the results. Under the aggressive $32\times$ scale factor, each coarse pixel corresponds to a $32\!\times\!32$ target block, and bicubic upsampling provides only limited structural guidance across block boundaries. As a result, all models exhibit a $7\!\times\!7$ checkerboard artifact in their error maps, although its severity varies substantially across architectures. SR3, EDSR, and SRGAN show the strongest error concentration along coarse-pixel boundaries, indicating greater difficulty in reconstructing coherent thermal structure across the upsampling grid. RCAN and the EFM-guided regression baseline produce more spatially uniform errors, but their overall error magnitude remains moderate. EFDiff-$\epsilon$ substantially suppresses these boundary artifacts relative to its HLS-concatenation counterpart SR3, indicating that EFM cross-attention conditioning helps resolve thermal transitions at coarse-pixel edges. EFDiff-$x_0$ achieves both the lowest error magnitude and the most spatially uniform distribution, effectively overcoming the checkerboard artifacts that persist in the other methods under this extreme super-resolution setting.

\subsection{Case Study.}
\label{sec:case_study}

Figure~\ref{fig:visual_comparison} presents a qualitative comparison on a representative test patch. SRGAN, SwinIR, HAT, and EDSR exhibit clear structural prediction errors visible in both the full-field and zoomed views: they produce only a coarse and often incorrect spatial distribution, failing to capture the underlying thermal structure. RCAN and the EFM-guided regression baseline improve upon this by recovering approximate spatial contours, but remain unable to reconstruct fine-scale thermal detail---their outputs represent blurred envelopes rather than resolved structure. SR3 and ResShift further sharpen the predicted contours, yet residual artifacts persist in both spatial positioning and absolute temperature values. The two EFM-conditioned models achieve the best reconstruction: both EFDiff-$\epsilon$ and EFDiff-$x_0$ correctly resolve the spatial contours of thermal boundaries and faithfully reproduce temperature discontinuities and absolute value differences that all other methods fail to capture. Between the two, EFDiff-$x_0$ achieves the lowest RMSE, while EFDiff-$\epsilon$ produces richer spatial texture at the cost of a slightly higher noise floor.

\subsection{Ablation and Analysis.}
\label{sec:ablation}

We conduct two ablations to validate the key design choices: the number of sampling steps and the effectiveness of EFM cross-attention conditioning.

\noindent \textbf{Sampling steps.}
Table~\ref{tab:ddim_ablation} examines how the number of reverse-process steps affects reconstruction quality on the test set. For EFDiff-$x_0$, the single-step variant already achieves the lowest RMSE (0.373$\,{}^{\circ}$C) and highest SSIM (0.743), indicating that the transition-based formulation concentrates most of the reconstruction into a very short trajectory. Increasing the number of steps up to the native 15-step schedule slightly worsens distortion metrics but improves perceptual quality, with LPIPS decreasing from 0.420 to 0.309. A non-monotonic pattern appears at intermediate step counts (3--5), where LPIPS is slightly worse than at 1 step, suggesting that partial multi-step refinement does not yet produce a perceptually coherent reconstruction. 
For EFDiff-$\epsilon$, LPIPS improves steadily as the number of DDIM steps increases, while RMSE and SSIM remain broadly stable before improving under the native 1{,}000-step schedule. The largest perceptual gain occurs in the first doubling from 25 to 50 steps, after which the improvement becomes more gradual. The native $1,000$-step schedule achieves the best performance, consistent with the variance-preserving formulation, which distributes the reconstruction across many timesteps and therefore benefits from a longer reverse trajectory.

\begin{table}[t]
\centering
\caption{Effect of sampling steps on reconstruction quality. Bold indicates the best result of each model; $\dagger$ denotes the native (non-DDIM) sampling schedule.}
\small
\setlength{\tabcolsep}{4pt}
\renewcommand{\arraystretch}{0.97}
\begin{tabular}{llccc}
\toprule
Model & Steps & RMSE (°C)$\downarrow$ & SSIM$\uparrow$ & LPIPS$\downarrow$ \\
\midrule
\multirow{5}{*}{EFDiff-$\epsilon$}
 & 25             & 0.518          & 0.626          & 0.330 \\
 & 50             & 0.520          & 0.624          & 0.306 \\
 & 100            & 0.526          & 0.617          & 0.294 \\
 & 250            & 0.531          & 0.611          & 0.286 \\
 & 1000$^\dagger$ & \textbf{0.498} & \textbf{0.642} & \textbf{0.223} \\
\midrule
\multirow{5}{*}{EFDiff-$x_0$}
 & 1              & \textbf{0.373} & \textbf{0.743} & 0.420 \\
 & 3              & 0.380          & 0.735          & 0.422 \\
 & 5              & 0.386          & 0.731          & 0.421 \\
 & 10             & 0.391          & 0.727          & 0.418 \\
 & 15$^\dagger$   & 0.439          & 0.699          & \textbf{0.309} \\
\bottomrule
\end{tabular}
\vspace{-0.4cm}
\label{tab:ddim_ablation}
\end{table}

\begin{figure}[t]
  \centering
  \includegraphics[width=\columnwidth]{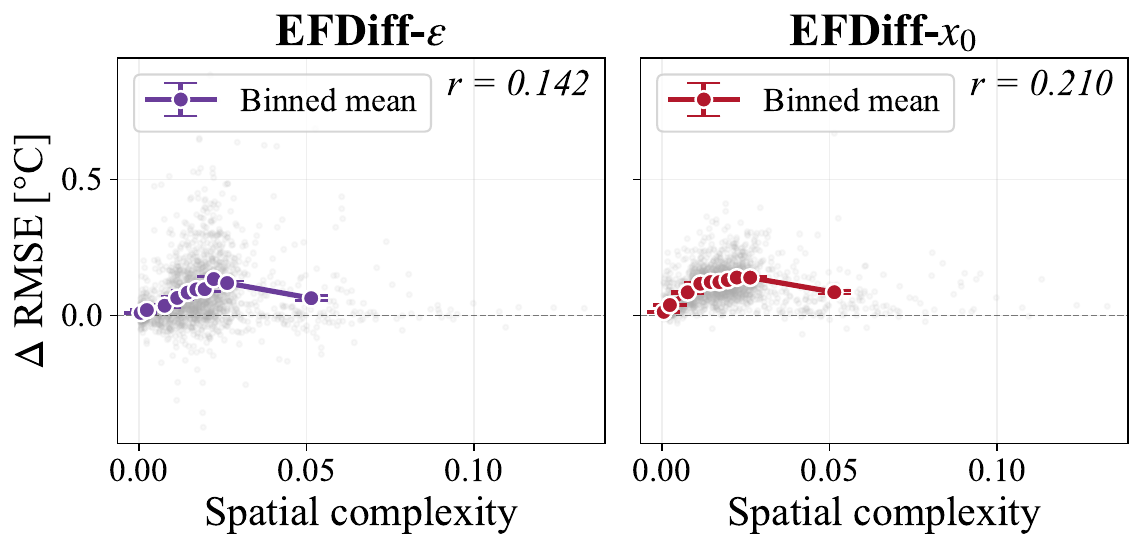}
  \caption{Per-patch $\Delta$RMSE versus spatial complexity for the two diffusion formulations. Grey dots denote individual patches; colored circles denote binned means with standard error bars.}
  \label{fig:efm_scatter}
  \vspace{-0.2cm}
\end{figure}

\noindent \textbf{Effectiveness of EFM conditioning.}
To understand where EFM cross-attention conditioning is most beneficial, we analyze the per-patch RMSE improvement,
$
\Delta\mathrm{RMSE} = \mathrm{RMSE}_{\mathrm{HLS}} - \mathrm{RMSE}_{\mathrm{EFM}},
$
as a function of scene complexity. We measure scene complexity by the mean gradient magnitude of the six HLS reflectance bands. Figure~\ref{fig:efm_scatter} shows this relationship for 2,000 randomly sampled test patches under both diffusion formulations.
Across both formulations, the binned means remain positive over nearly the entire complexity range, and the Pearson correlation is positive in both panels. This indicates that EFM conditioning consistently outperforms HLS channel concatenation, with the advantage becoming larger as scene complexity increases. The gain rises rapidly from low-complexity scenes, reaches its maximum near a complexity value of 0.025, and then decreases slightly while remaining positive over the 0.025--0.05 range. Beyond 0.05, the number of samples becomes small, making the binned statistics less stable; however, most individual patches still lie above zero, suggesting that the EFM advantage persists. In contrast, when the complexity is below about 0.025, $\Delta\mathrm{RMSE}$ stays close to zero, indicating that the two conditioning strategies perform similarly because the coarse thermal input already constrains the reconstruction well. These results show that EFM embeddings are most helpful in spatially heterogeneous scenes, where raw spectral bands alone provide insufficient semantic information to resolve fine-scale thermal ambiguity.

\begin{figure}[t]
  \centering
  \includegraphics[width=\columnwidth]{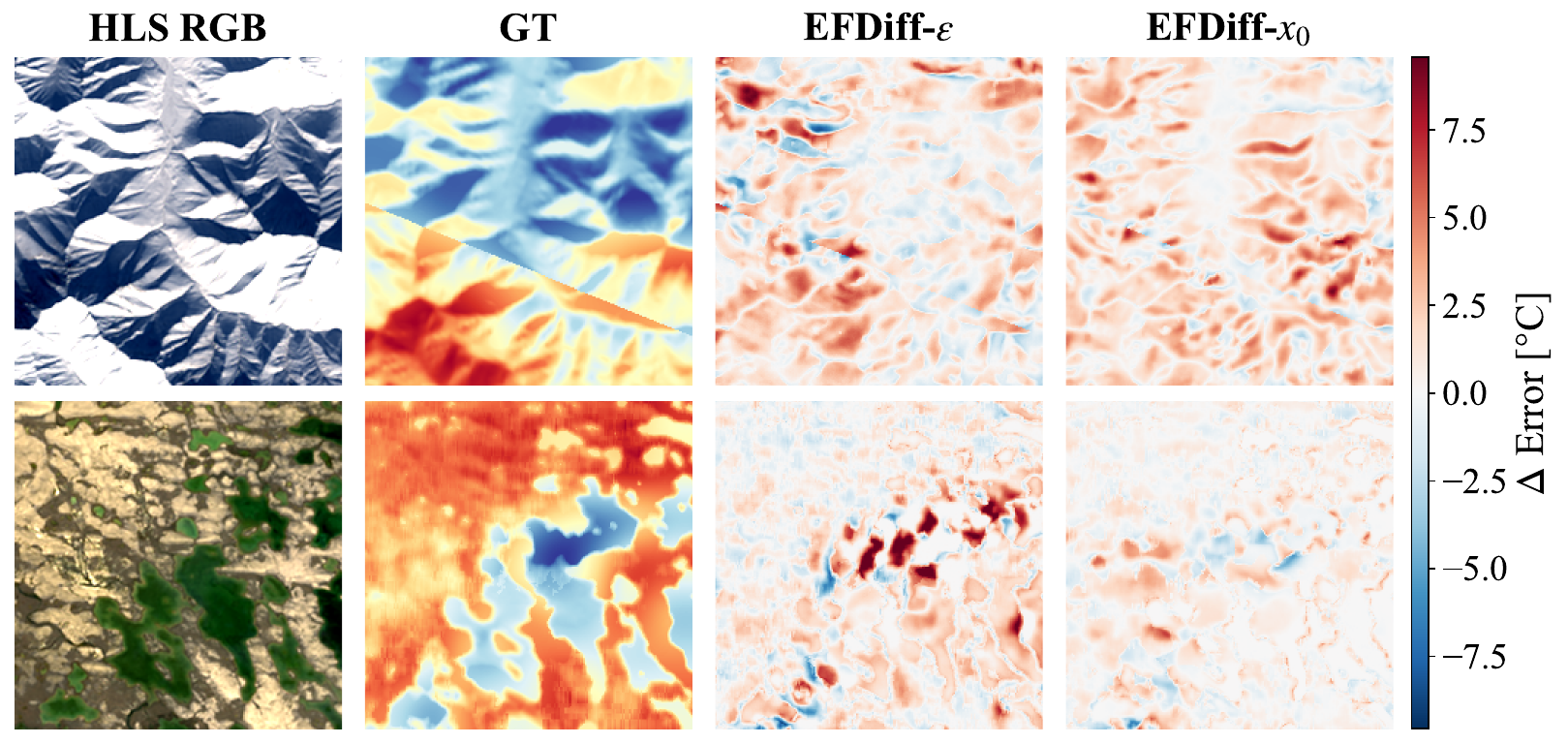}
  \caption{Error difference maps for two selected test patches. Red indicates lower absolute error with EFM conditioning, and blue indicates lower absolute error with HLS channel concatenation.}
  \label{fig:efm_error_maps}
  \vspace{-0.4cm}
\end{figure}

Figure~\ref{fig:efm_error_maps} provides a spatial view of this effect for two test patches with large EFM gains. Each error difference map shows
$
|\mathrm{error}_{\mathrm{HLS}}| - |\mathrm{error}_{\mathrm{EFM}}|,
$
so red indicates locations where EFM conditioning reduces absolute error, while blue indicates locations where HLS channel concatenation performs better. In both examples, the maps are dominated by red regions, and the largest gains are concentrated along land-cover boundaries visible in the HLS RGB images. These are precisely the locations where a $32\times$ coarse pixel mixes multiple thermal classes, making the conditioning signal especially important for resolving fine-scale structure. Blue regions appear only sporadically and are mostly confined to spectrally homogeneous interiors, where both conditioning strategies already perform well, and the remaining differences are due to sampling variability.

\section{Conclusion.}
\label{sec:conclusion}

In this paper, we introduced EFDiff and a novel methodological perspective: when EFMs meet diffusion, pretrained geospatial semantics can serve as a strong prior for generative reconstruction. On a globally diverse benchmark for extreme $32\times$ LST super-resolution, EFDiff consistently outperformed deterministic, adversarial, and diffusion baselines, while EFDiff-$x_0$ and EFDiff-$\epsilon$ offered complementary advantages in fidelity and perceptual realism. More broadly, our results suggest that pretrained Earth representations and diffusion are complementary: the former provide transferable spatial semantics, and the latter translate them into plausible fine-scale predictions. Although we presented EFDiff in the context of LST super-resolution, the framework is broadly applicable to remote sensing problems where pretrained geospatial representations can guide generative reconstruction. We hope this work motivates broader task-specific and application-driven instantiations of this paradigm across remote sensing and other scientific domains.

\bibliographystyle{siamplain}
\bibliography{references}

\appendix

\section{Dataset Construction Details.}
\label{sec:appendix_data}

Our benchmark is constructed from NASA's \texttt{HLSL30}~v2.0 product~\cite{neigh2021hlsl30}, which provides atmospherically corrected 30\,m Landsat~8/9 reflectance in six bands (Blue, Green, Red, NIR, SWIR1, and SWIR2) together with the \texttt{Fmask} quality layer. Thermal targets are taken from the Landsat Collection~2 Level-2~\cite{usgs_landsat_c2_l2} \texttt{ST\_B10} band~\cite{cook2014development}. We convert \texttt{ST\_B10} to Celsius using the standard scaling coefficients provided in the Landsat~8--9 Collection~2 Level-2 Science Product Guide~\cite{usgs2024landsatl2spguide}. Although the native TIRS thermal resolution is 100\,m, the Level-2 product is distributed on a 30\,m grid. To ensure pixel-level alignment between reflectance and thermal observations, we request both products on an identical grid through Google Earth Engine~\cite{gorelick2017google}. Samples without a valid LST match within $\pm$1\,day are discarded.

To promote global diversity, we adopt the tile sampling strategy introduced in Prithvi-EO-2.0~\cite{szwarcman2025prithvi}. Specifically, tiles are selected using LULC-proportional sampling with urban oversampling and entropy-based gap filling, based on the Copernicus 100\,m global land-cover product~\cite{buchhorn2020copernicus} and the RESOLVE 2017 ecoregion dataset~\cite{dinerstein2017ecoregion}. Tiles from Antarctica and central Greenland are removed due to limited relevance and sparse valid thermal coverage. The remaining 3{,}094 MGRS tiles are then split geographically into 2{,}943 training tiles and 151 test tiles.
For each selected tile, we retrieve up to 48 observations from 2014 to 2026, subject to a scene-level cloud threshold of less than 20\%. Each observation is partitioned into non-overlapping $256\!\times\!256$ patches. We then apply patch-level quality filtering based on cloud coverage, invalid pixels, and missing thermal values. To prevent spatial leakage, any training patch overlapping the footprint of a test tile is excluded. In addition, highly homogeneous ocean and desert patches are subsampled to reduce class imbalance and avoid over-representing visually trivial regions.

During model training and evaluation, each $256\!\times\!256$ patch is converted into a $224\!\times\!224$ sample. We use random cropping during training and deterministic center cropping during inference. The LR thermal input is synthesized according to Wald's reduced-resolution protocol~\cite{wald1997fusion}. Concretely, we first apply a Gaussian point spread function with $\sigma = s/\pi$, where $s=32$ is the scale factor, to simulate sensor blur. We then perform area-average downsampling to obtain a $7\!\times\!7$ coarse thermal field, which is bicubically upsampled back to $224\!\times\!224$ to form $\tilde{X}$.

The prediction target is the residual
$
R = Y - \tilde{X}.
$
We normalize reflectance bands to $[0,1]$. Residuals are standardized as
$
R' = \frac{R - \mu_R}{3\sigma_R},
$
where $\mu_R$ and $\sigma_R$ are computed from the training set. This scaling maps approximately 99.7\% of residual values into the interval $[-1,1]$, which stabilizes diffusion training while preserving the physical dynamic range of the reconstructed signal.

\section{Implementation and Training Details.}
\label{sec:appendix_setup}

\begin{table}[t]
\centering
\caption{Dataset statistics by land cover category.}
\label{tab:dataset_stats}
\small
\setlength{\tabcolsep}{4pt}
\renewcommand{\arraystretch}{0.98}
\begin{tabular}{lrrrrrr}
\toprule
& \multicolumn{2}{c}{Tiles} & \multicolumn{2}{c}{Patches} \\
\cmidrule(lr){2-3} \cmidrule(lr){4-5}
Category & Train & Test & Train & Test \\
\midrule
Forest         &   850 &  43 &  62{,}541 &  3{,}300  \\
Low Vegetation & 1{,}513 &  72 & 121{,}709 &  5{,}741 \\
Water          &   580 &  38 &  46{,}409 &  2{,}897 \\
\midrule
\textbf{Total} & \textbf{2{,}943} & \textbf{153} & \textbf{230{,}659} & \textbf{11{,}938} \\
\bottomrule
\end{tabular}
\vspace{-0.5cm}
\end{table}

All experiments are conducted on $224\!\times\!224$ patches under the $32\times$ super-resolution protocol described in Sec.~\ref{sec:dataset}. All models predict the residual $R = Y - \tilde{X}$ rather than the HR field directly.

The proposed EFDiff-$\epsilon$ and EFDiff-$x_0$ models share the same denoising backbone described in Sec.~\ref{sec:network}. The UNet uses inner channel 64, channel multipliers $(1,2,4,8,16)$, two residual blocks per stage, and self-/cross-attention at the $14\!\times\!14$ resolution, with GroupNorm-16 throughout. Geospatial conditioning is provided by a frozen Prithvi-EO-2.0-300M encoder, which outputs 196 spatial embeddings of dimension $1024$. These embeddings are injected into the denoiser through cross-attention with 8 attention heads.

For EFDiff-$\epsilon$, we use a variance-preserving linear noise schedule with $T=1{,}000$ and $\beta_t$ linearly increasing from $10^{-6}$ to $10^{-2}$. The model is trained with an L1 objective and learning rate $2\times 10^{-5}$. At inference, we use DDIM sampling with 50 reverse steps. For EFDiff-$x_0$, we adopt the ResShift exponential transition schedule with $T=15$ and $\kappa=1.0$, train with an L2 objective, and use learning rate $5\times 10^{-5}$.

We compare against nine baselines. Among deterministic residual-learning models, EDSR~\cite{lim2017enhanced} uses 64 feature channels and 16 residual blocks, while RCAN~\cite{zhang2018image} uses 10 residual groups with 20 blocks per group and channel attention. SRGAN~\cite{ledig2017photo} follows the SRResNet generator with VGG-19 perceptual loss and the LSGAN objective. SwinIR~\cite{liang2021swinir} and HAT~\cite{chen2023activating} are configured with embedding dimension 60 and window size 7. The diffusion baselines SR3~\cite{saharia2022image} and ResShift~\cite{yue2023resshift} share the same UNet backbone as the proposed models but use HLS channel concatenation instead of EFM cross-attention conditioning. We also include an EFM-guided regression baseline that uses the same frozen Prithvi encoder as the proposed method but predicts the residual in a single forward pass. All learned baselines are trained with L1 loss.
All models are optimized with Adam using gradient clipping at 1.0. Training is performed on 2$\times$H200 GPUs with \texttt{bf16} precision. Deterministic and adversarial baselines are trained for 100k iterations with a learning rate of $10^ {-4}$, while diffusion models are trained for 200k iterations. For all methods, we use EMA with decay 0.9999 for evaluation.

We evaluate reconstruction quality along two complementary axes. Distortion metrics include RMSE in ${}^{\circ}$C and SSIM~\cite{wang2004image}, which measure pixel-level fidelity and structural similarity to the ground truth. Perceptual metrics include LPIPS~\cite{zhang2018unreasonable} and FID~\cite{heusel2017gans}, which assess whether generated thermal fields exhibit realistic local texture and match the overall distribution of real samples. Because extreme $32\times$ super-resolution is highly ill-posed, we report both classes of metrics to capture the perception--distortion trade-off~\cite{blau2018perception}. To further expose performance differences across terrain types, all metrics are reported separately for three land-use/land-cover groups derived from the Copernicus 100\,m land-cover product~\cite{buchhorn2020copernicus}: \emph{Forest} (closed and open forest), \emph{Low Vegetation} (managed vegetation, herbaceous vegetation, shrubs, bare vegetation, and moss), and \emph{Water} (sea, permanent water, and wetland).

\end{document}